%% file: main.tex
\pdfoutput=1

\documentclass[11pt]{article}

\usepackage[preprint]{acl}

\usepackage{times}
\usepackage{latexsym}
\usepackage{graphicx}
\usepackage{multirow}
\usepackage{multicol}
\usepackage{amsmath}
\usepackage{amsfonts}
\usepackage{amssymb}
\usepackage{algorithm}
\usepackage{algpseudocode}
\usepackage{stix}
\usepackage{booktabs}
\usepackage{xcolor}

\newcommand{\argmin}{\mathop{\mathrm{argmin}}}

\usepackage[T1]{fontenc}

\usepackage[utf8]{inputenc}

\usepackage{microtype}

\usepackage{inconsolata}

\title{Multi-Dimensional Optimization for Text Summarization via Reinforcement Learning}

 \author{Sangwon Ryu\thanks{Equal contribution}$^{1}$, Heejin Do\footnotemark[1]$^{1}$, Yunsu Kim$^{3}$, Gary Geunbae Lee$^{1,2}$, Jungseul Ok$^{1,2}$ \\
  \centering
  \begin{tabular}[t]{c}
    $^{1}$Graduate School of Artificial Intelligence, POSTECH, South Korea \\
    $^{2}$Department of Computer Science and Engineering, POSTECH, South Korea \\
    $^{3}$aiXplain Inc., Los Gatos, CA, USA \\
    \texttt{\{ryusangwon, heejindo, gblee, jungseul\}@postech.ac.kr}, \texttt{yunsu.kim@aixplain.com} \\
  \end{tabular}
}

\begin{document}
\maketitle


\input{content/abstract}

\input{content/intro}

\input{content/related}

\input{content/method}

\input{content/experiment}

\input{content/result}

\input{content/analysis}

\input{content/future}

\input{content/conclusion}

\input{content/limitation}
\input{content/ethics}

\input{content/acknowledge}

\bibliography{custom}

\input{content/appendix}

\end{document}

%% file: content/abstract.tex


\begin{abstract}
The evaluation of summary quality encompasses diverse dimensions such as \textit{consistency}, \textit{coherence}, \textit{relevance}, and \textit{fluency}. However, existing summarization methods often target a specific dimension, facing challenges in generating well-balanced summaries across multiple dimensions. In this paper, we propose multi-objective reinforcement learning tailored to generate balanced summaries across all four dimensions. We introduce two multi-dimensional optimization (MDO) strategies for adaptive learning: 1) MDO$_{\text{min}}$, rewarding the current lowest dimension score, and 2) MDO$_{\text{pro}}$, optimizing multiple dimensions similar to multi-task learning, resolves conflicting gradients across dimensions through gradient projection. Unlike prior ROUGE-based rewards relying on reference summaries, we use a QA-based reward model that aligns with human preferences. Further, we discover the capability to regulate the length of summaries by adjusting the discount factor, seeking the generation of concise yet informative summaries that encapsulate crucial points. Our approach achieved substantial performance gains compared to baseline models on representative summarization datasets, particularly in the overlooked dimensions. 
\end{abstract}

%


%% file: content/intro.tex
\section{Introduction}

\input{fig/fig-example}

Determining a "good summary" extends beyond a single factor, generally embracing multiple dimensions such as \textit{coherence}, \textit{consistency}, \textit{fluency}, and \textit{relevance} \cite{kryscinski2019neural, zhong-etal-2022-towards, liu2022reference, wang2023element, liu-etal-2023-g}. Despite the remarkable advancements in abstractive summarization, challenges persist in addressing issues such as factual inconsistency, which generates inaccurate information, and irrelevance, which involves omitting crucial details.

Recently, there have been ongoing efforts to focus on such inferior dimensions \cite{pasunuru-bansal-2018-multi, gunasekara-etal-2021-using-question, cao-etal-2022-hallucinated, berezin-batura-2022-named, wan2023faithfulnessaware, liu2023improving, nan-etal-2021-entity, wang2023element, chern-etal-2023-improving}, and reinforcement learning (RL) is applied as one strategy. Most existing RL approaches use a single reward of the ROUGE score \cite{lin-2004-rouge}, which measures the overlap with the reference summary. However, its subpar quality across various datasets has been frequently underscored \cite{liu2023learning, zhang2023benchmarking, goyal2023news}.

Pointing out the limitations of ROUGE scores in detecting hallucinations, various studies have focused on addressing this issue. \citet{pasunuru-bansal-2018-multi} assigned weights to each word to overcome shortcomings of ROUGE, \citet{roit-etal-2023-factually} provided a reward with the natural language inference (NLI) entailment relationship between generated summary and the document, and \citet{gunasekara-etal-2021-using-question} provided rewards via Question Answering (QA) model. However, those methods cannot capture summary-intrinsic dimensions, such as \textit{fluency} or \textit{coherence}. Addressing shortcomings in one dimension often leads to unintended drawbacks in other dimensions; thus, achieving a high-quality summary generation by balancing multiple dimensions remains challenging (Figure~\ref{fig: figure 1}).


In this work, we introduce multi-objective RL, aiming to generate solid summaries that are coherent, factually consistent, fluent, and relevant. Our RL approach is based on a proximal policy optimization (PPO) \cite{schulman2017proximal}, and we incorporate four dimensions of a unified multi-dimensional evaluation metric, \texttt{UniEval} \cite{zhong-etal-2022-towards}, as multiple rewards. We suggest two strategies for optimal rewarding with multiple objectives, namely MDO$_\text{min}$ and MDO$_\text{pro}$. MDO$_\text{min}$ fosters adaptive learning by selecting the lowest dimension score as the reward at each iteration. Meanwhile, MDO$_\text{pro}$ projects gradient onto the normal plane to handle conflicting gradients in multi-task RL, leveraging a PCGrad \cite{yu2020gradient} optimizer. By effectively projecting the gradients of multiple rewards, our method can adjust the learning direction for optimal training. Both strategies aim to enhance deficient dimensions while preserving superior ones during training.

In summarization tasks, unlike typical PPO usage that rewards at each step, the score for a generated summary is obtained only at the end of the episode when the entire summary is produced. KL-penalty replaces the reward per token during episodes; hence, the discount factor can be crucial in obtaining an optimal policy \cite{kim2022adaptive}. Consequently, we investigate how adjusting the discount factor affects the generated summaries, particularly in length.

Our MDO strategies outperform the baseline model in experiments using the representative CNN/DM and BillSum summarization datasets. Notably, our methods significantly enhance the previously inferior \textit{relevance} dimension, supporting competitive results in other dimensions. Additional examinations, measuring whether the contents of the generated summaries are from the original articles, reveal around 90\% coverage with a shorter average length. This outcome implies the capacity of the MDO to create brief yet pertinent summaries. 


Our contributions are summarized as follows: 
\begin{itemize}
    \item We propose two multi-dimensional optimization methods for multi-objective RL, introducing multiple \texttt{UniEval} dimensions as rewards.
    \item We have empirically verified improvements in deficient dimensions while maintaining competitiveness in superior dimensions across two datasets, outperforming naive MDO methods.
    \item We find that adjusting a discount factor can control the generated summary length.
\end{itemize}


\input{fig/fig-main}



%% file: fig/fig-example.tex
\begin{figure}[]
\centering
\includegraphics[width=0.48\textwidth]{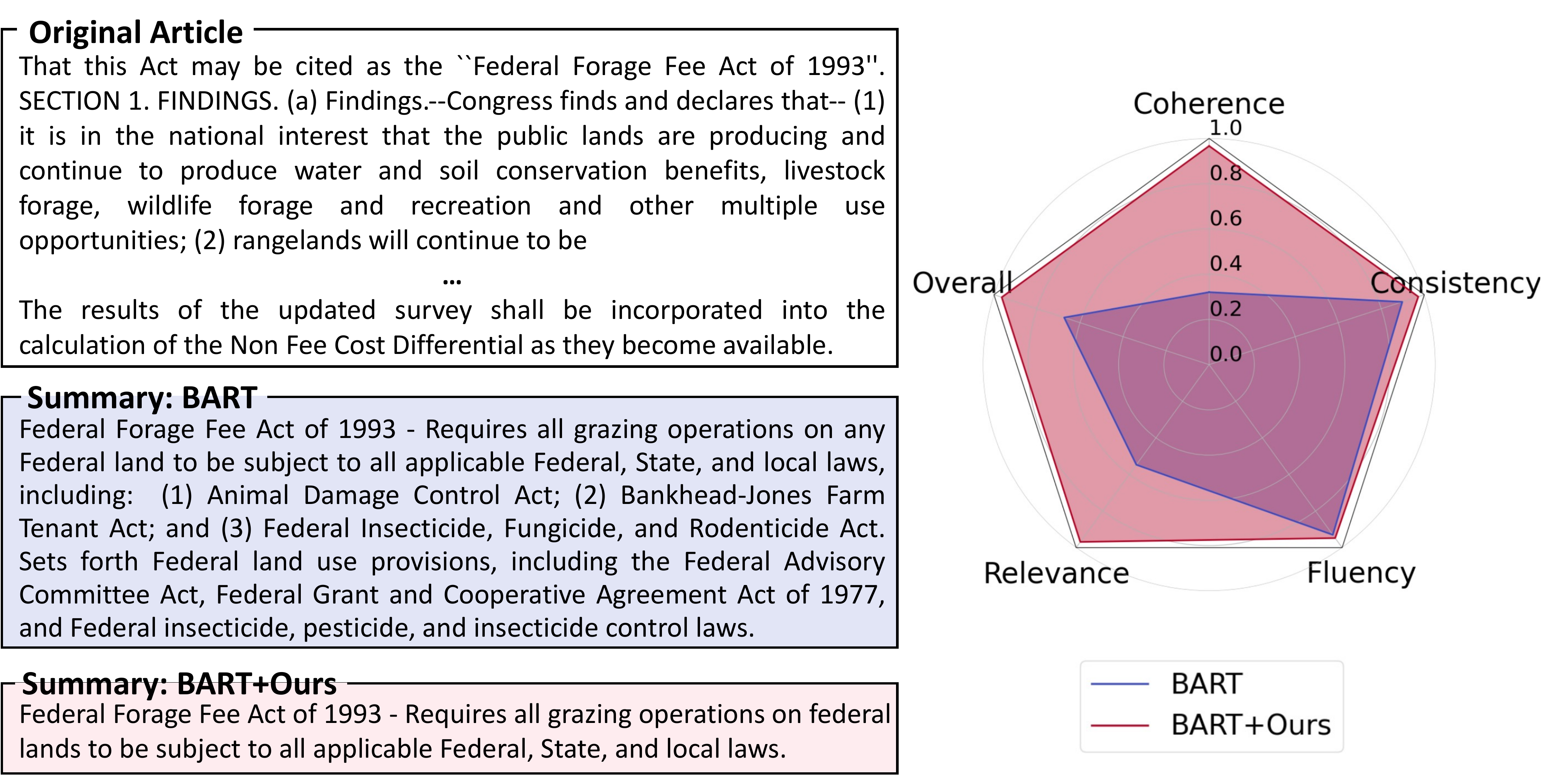}
\caption{While the baseline model produces an imbalanced summary ({$\color{blue!45}\mdblksquare$}), we aim to generate overall high-quality summaries ({$\color{red!45}\mdblksquare$}). The radar chart illustrates \texttt{UniEval} scores for four dimensions.}
\label{fig: figure 1}
\end{figure}



%% file: fig/fig-main.tex
\begin{figure*}[]
\centering
\includegraphics[width=0.93\textwidth]{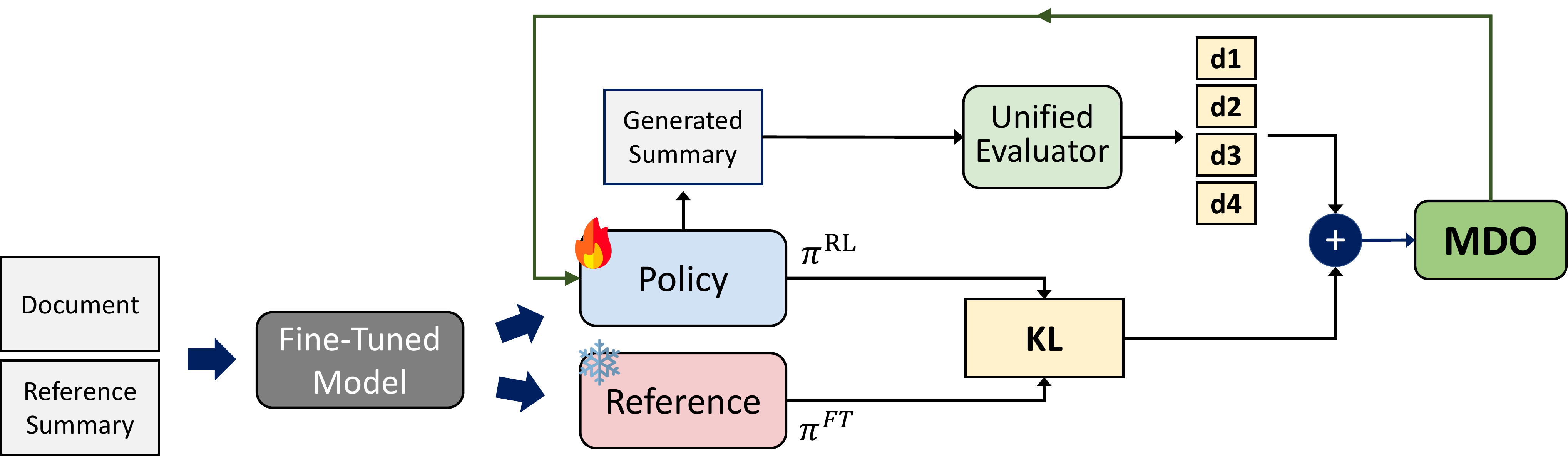}
\caption{Entire process of Multi-dimensional Optimization (MDO). Through MDO, we optimize the scores for each dimension while training the policy. $d1$, $d2$, $d3$, and $d4$ refer to \textit{coherence}, \textit{consistency}, \textit{fluency}, and \textit{relevance}, respectively.}
\label{fig: main}
\end{figure*}

%% file: content/related.tex
\section{Related Work}

\paragraph{Dimension-specific text summarization} Previous studies have mostly focused on improving specific dimensions of text summarization, such as generating consistent summaries by resolving hallucinations. \citet{wang2023element} involves a two-stage process where key entities are first extracted in the initial stage, followed by the integration of these entities to generate summaries in the second stage. \citet{wan2023faithfulnessaware} altered the decoding strategy using a ranker and lookahead approach to produce the token with the highest faithfulness score. Their methods only considers to generate faithful summaries but overlooks other various dimensions.



\paragraph{RL for abstractive text summarization} RL methods for text summarization have predominantly utilized the ROUGE score as a reward \cite{narayan-etal-2018-ranking, chen2018fast, pasunuru-bansal-2018-multi, kryscinski-etal-2018-improving, dong-etal-2018-banditsum, paulus2017deep, wang2020reinforced, parnell-etal-2022-multi}. However, recent studies emphasized that the ROUGE score fails to evaluate summaries adequately due to the revealed poor quality of reference summaries in summarization tasks \cite{liu2023learning, zhang2023benchmarking, goyal2023news}. Moreover, the ROUGE score only calculates the word overlap with the reference summary, failing to evaluate whether sentences are natural or consistent.

Therefore, some researchers have explored the application of the NLI model \cite{roit-etal-2023-factually} or QA model \cite{gunasekara-etal-2021-using-question} as a reward, which does not solely rely on the ROUGE score. \citet{roit-etal-2023-factually} employs reinforcement learning with an NLI reward, aiming to maintain high \textit{consistency} by using the entailment relationship between the summary and the document as a reward. \citet{gunasekara-etal-2021-using-question} generate questions from both the document and the summaries using a QA model to verify the presence of answers, aiming to enhance precision and recall related to \textit{consistency} and \textit{relevance}. Yet, these methods do not comprehensively consider diverse quality dimensions.

\paragraph{Multi-objective RL} RL with multiple rewards can lead to more efficient model training \cite{pmlr-v202-dann23a}. However, multi-reward application in text summarization has not been extensively explored. \citet{pasunuru-bansal-2018-multi} employ multiple rewards such as ROUGE-L, ROUGE-Sal (which weighs vital information), and entailment, but they simply approach as multi-task learning without consideration for finer optimization. \citet{9868151} utilize multiple RL policies to summarize multiple documents by constructing individual policy models for \textit{importance}, \textit{redundancy}, and \textit{length}. They aim to concisely summarize multiple documents, preventing content overlap and including only salient information. Yet, they did not aim for a comprehensive summary of a single document, as only the \textit{importance} feature was considered. Unlike their exclusive focus on enabling the model to capture the essential or relevant content, we explore the optimal strategies for multi-objective RL, aiming for well-balanced summarization.


%% file: content/method.tex
\section{Method Description}

Throughout the RL process, it is crucial to maintain the fundamental summarization capabilities of the fine-tuned model while simultaneously improving scores across various dimensions. To achieve this goal, we employ proximal policy optimization (PPO) \cite{schulman2017proximal} for RL application, utilizing a supervised, parameter-frozen reference model to guide the policy. In our pursuit of multi-objective RL in summarization, we adopt \texttt{UniEval} \cite{zhong-etal-2022-towards}, a metric that evaluates scores across different dimensions using a QA model. Incorporating four dimensions in the rewarding process, we introduce two optimal MDO methods to guide RL policy updates effectively. The entire process is illustrated in Figure \ref{fig: main}.

\paragraph{Multi-rewards} \texttt{UniEval} leverages a QA module for a unified multi-dimensional assessment in the rewarding process. The dimensions tackled by the \texttt{UniEval} closely align with human preferences, evaluating summaries based on key quality indicators. They include \textit{coherence} (the structural coherence of the summary), \textit{consistency} (the absence of discrepancies with the main text), \textit{fluency} (the natural flow of sentences within the summary), and \textit{relevance} (the inclusion of only important content from the document).




\paragraph{PPO}
PPO stands out as a well-established policy gradient model, renowned for its efficiency and stability attributed to its clipping surrogate objective. This object mitigates abrupt changes during policy updates, ensuring overall stability and avoiding divergence. Given the clipped surrogate objective, $L^{CL}$, the value loss, $L^{VF}$, and the entropy, $S$, the full PPO loss at timestep $t$ is defined as follows:
\[
\mathrm{L}_t(\theta) = \hat{\mathbb{E}}_t \Bigl[\mathrm{L}^{\mathrm{CL}}_t(\theta) \notag - c_1 \mathrm{L}^{\mathrm{VF}}_t (\theta) + c_2 S[\pi_\theta](s_t) \Bigr]
\]

Unlike typical PPO applications that provide rewards at each time step, the summary can only be evaluated once when the entire sentences are generated in the summarization task. Thus, in line with the approach proposed by \citet{stiennon2022learning}, we employ a supervised fine-tuned summarization model as the policy $\pi^{\mathrm{RL}}$. The value model shares parameters with $\pi^{\mathrm{RL}}$, with an additional value head. Again, we utilize a reference model $\pi^{\mathrm{FT}}$, which is also a fine-tuned summarization model but with frozen parameters, to maintain the summarization performance of the $\pi^{\mathrm{RL}}$. In particular, rewards for each action, except for the generation of the last token, is the KL penalty between the policy $\pi^{RL}$ and the reference model $\pi^{FT}$. This process ensures that the $\pi^{RL}$ does not diverge too far from the supervised fine-tuned summarization model during the RL process. For the final action, which is the selection of the last token of the summary, a total reward is assigned by a reward model, $r(x, y)$, for the entire summary:
\[
\mathrm{R}(x,y) = r(x, y) - \beta \log[\pi_{\theta}^{\mathrm{RL}}(y|x)/\pi^{\mathrm{FT}}(y|x)]
\]

Generalized advantage estimation (GAE) \cite{schulman2018highdimensional} is used for advantage estimation. Finely adjusting the influence of future reward in GAE is facilitated by employing the discount factor $\gamma$ alongside parameter $\lambda$ . $x$ and $y$ denote the document and summary, respectively. The state $s$ is the current token, the action $a$ is the selection of the next token by the $\pi^{RL}$, and the action space is the vocabulary of the $\pi^{RL}$, $V$. 

In our multi-objective setting, the score for each dimension $d_{k}$ corresponds to a reward $r_{k}(x, y)$. The key focus of our two MDO strategies lies in optimizing these multi-rewards to train the policy effectively. We use online learning, similar to the previous methods \cite{stiennon2022learning}, which demonstrated strong performance across various domains \cite{NEURIPS2023_fc65fab8}.



\input{fig/algo-mdomin}

\subsection{MDO$_{\text{min}}$} Focusing on the most vulnerable dimensions, we suggest MDO$_{\text{min}}$, which selects a minimum dimension score as the reward, $r(x, y)$, among the evaluated four-dimensional scores. This approach intuitively aims to uplift the performance of the inferior-quality dimensions. By adopting the minimum score, the model is prompted to perform policy gradients to address the weakest dimension, achieving a balanced summary generation. The same model evaluates all four dimensions; thus, no scaling is required, and the lowest-rated dimension is directly utilized as the reward. The details of the MDO$_{\text{min}}$ is explained in Algorithm \ref{algo: mdo_min}.

\input{fig/algo-mdopro}

\subsection{MDO$_{\text{pro}}$} While rewards can be adaptively provided based on individual dimension scores, it may prove insufficient if there exists an inherent trade-off relationship between dimensions. For instance, attempting to improve \textit{consistency} by including entities from the main document in the summary could potentially reduce the \textit{fluency} between sentences within the summary. Consequently, finding a Pareto improvement becomes challenging when faced with such inherent trade-offs.


\input{fig/tab-main}

\input{fig/tab-cnn}

To overcome the intrinsic trade-off relationship, we suggest an MDO$_{pro}$, which projects multiple conflicting gradients onto a plane, utilizing the PCGrad optimizer \cite{yu2020gradient}. Treating multiple dimensions as distinct tasks, the optimizer projects each task's gradient onto the normal plane of the gradient of other tasks with conflicting gradients. In cases where gradients from multiple losses oppose each other, the learning may become ineffective. The PCGrad optimizer alleviates interference between the gradients of different dimensions by ensuring that the gradient of one dimension does not adversely affect the gradient of others. The detailed process is outlined in Algorithm \ref{algo: mdo_pro}. 



%% file: fig/algo-mdomin.tex
\begin{algorithm}[h]
\small
\renewcommand{\algorithmicrequire}{\textbf{Input:}}
\caption{Calculation of MDO$_{\text{min}}$}
\label{algo: mdo_min}
\begin{algorithmic}[1]
\Require documents=$\{ D_1, D_2, \ldots, D_\mathcal{N} \}$,\\
policy $\pi_{\theta}$, model parameter $\theta$, $Evaluator$ \\
hyperparameter $\beta, \lambda$, discount factor $\gamma$,
\State $ Dims \gets \{\text{``coh''}, \text{``con''}, \text{``flu''}, \text{``rel''}\}$
\State $ \mathcal{M} \gets length(Dims)$
\For{$i=1$ \textbf{to} $\mathcal{N}$}
    \State $L \gets 0$
    \State \textit{// Generate a summary}
    \State $S_{i}$ = $\pi_{\theta}(D_{i})$
    \State \textit{// Calculate rewards}
    \For{$j = 1$ \textbf{to} $\mathcal{M}$}
        \State $r_{j} = Evaluator(Dims[j])$
    \EndFor
    \State $r = \argmin_{1 \leq m \leq \mathcal{M}} r_{m}(D_i, S_i)$
    \State $R = r(D_i, S_i) - \beta \log\left(\frac{\pi_{\theta}^{RL}(S_i|D_i)}{\pi^{FT}(S_i|D_i)}\right)$
    \State \textit{// Estimate advantage $\hat{A}$ using GAE}
    \State $\delta \gets r_t + \gamma V(s_{t+1}) - V(s_t)$
    \State $\hat{A_t} \gets \delta_t + \gamma \lambda \delta_{t+1} + \cdots + (\gamma \lambda)^{T-t+1} \delta_{T-1}$
    \State $L \gets \text{PPO loss for } \hat{A_t}, R, \pi_{\theta}$
    \State update $\Delta \theta$

\EndFor
\end{algorithmic}
\end{algorithm}

%% file: fig/algo-mdopro.tex
\begin{algorithm}[h]
\small
\renewcommand{\algorithmicrequire}{\textbf{Input:}}
\caption{Calculation of MDO$_{\text{pro}}$}
\label{algo: mdo_pro}
\begin{algorithmic}[1]
\Require documents=$\{ D_1, D_2, \ldots, D_\mathcal{N} \}$,\\
policy $\pi_{\theta}$, model parameter $\theta$, $Evaluator$ \\
hyperparameter $\beta, \lambda$, discount factor $\gamma$,
\State $ Dims \gets \{\text{``coh''}, \text{``con''}, \text{``flu''}, \text{``rel''}\}$
\State $ \mathcal{M} \gets \text{length}(Dims)$
\For{$i=1$ \textbf{to} $\mathcal{N}$}
    \State $L \gets 0$
    \State \textit{// Generate a summary}
    \State $S_{i}$ = $\pi_{\theta}(D_{i})$
    \State \textit{// Calculate rewards}
    \For{$j = 1$ \textbf{to} $\mathcal{M}$}
        \State $r_{j} = Evaluator(Dims[j])$
        \State $R_{j} = r_{j}(D_i, S_i) - \beta \log\left(\frac{\pi_{\theta}^{RL}(S_i|D_i)}{\pi^{FT}(S_i|D_i)}\right)$
        \State \textit{// Estimate advantage $\hat{A}$ using GAE}
        \State $\delta \gets r_t + \gamma V(s_{t+1}) - V(s_t)$
        \State $\hat{A_t} \gets \delta_t + \gamma \lambda \delta_{t+1} + \cdots + (\gamma \lambda)^{T-t+1} \delta_{T-1}$
        \State $L \gets L + \text{PPO loss for } \hat{A_t}, R_{j}, \pi_{\theta}$
    \EndFor

    \State $g_{m} \gets \nabla_{\theta}L(\theta) \; \forall m \in Dims$
    \State $g^{PC}_m \gets g_{m} \; \forall m$
    \State \textit{// Project conflict gradient}
    \State $(p, q) \gets \text{select } (p, q) \in Dims \times Dims \text{ where } p \neq q$
    \If{$g^{PC}_{p} \cdot g_{q} < 0$}
        \State $g^{PC}_{p} \gets g^{PC}_{p} - \frac{g^{PC}_{p} \cdot g_{q}}{\|g_{q}\|^2} g_{q}$
    \EndIf
    \State update $\Delta \theta = g^{PC} = \sum_{m} g^{PC}_{m}$

\EndFor
\end{algorithmic}
\end{algorithm}

%% file: fig/tab-main.tex
 


\begin{table*}
\centering
\scalebox{0.78}
{\begin{tabular}{l|c|ccccc|cc}
\toprule
\multicolumn{2}{l|}{} & \multicolumn{5}{c|}{UniEval} & \multicolumn{2}{c}{} \\ \cline{1-9}
{Model} & {Fine-tune} & Coherence & Consistency & Fluency & Relevance & Overall & QuestEval & BERT{\small Score} \\ 
\midrule
PEGASUS & SFT & 0.823 & 0.832 & 0.849 & 0.814 & 0.830 & 0.392 & 0.899 \\
\midrule
BART$_{base}$  & SFT & 0.838 & 0.833 & 0.845 & 0.779 & 0.824 & 0.425 & 0.902 \\ 
BART$_{base}$ & SFT+MDO$_{\text{min}}$ & \textbf{0.859} & \textbf{0.857} & \textbf{0.853} & \underline{0.806} & \textbf{0.843} & \textbf{0.431} & \textbf{0.924} \\
BART$_{base}$ & SFT+MDO$_{\text{pro}}$  & \underline{0.857} & \underline{0.853} & \underline{0.846} & \textbf{0.813} & \underline{0.842} & \underline{0.428} & \textbf{0.924} \\
\midrule
BART$_{large}$ & SFT & 0.884 & 0.865 & 0.864 & 0.843 & 0.864 & 0.424 & 0.904 \\ 
BART$_{large}$ & SFT+MDO$_{\text{min}}$  & \underline{0.899} & \underline{0.894} & \textbf{0.882} & \underline{0.869} & \textbf{0.886} & \textbf{0.435} & \textbf{0.924} \\
BART$_{large}$ & SFT+MDO$_{\text{pro}}$  & \textbf{0.900} & \textbf{0.895} & \underline{0.877} & \textbf{0.871} & \textbf{0.886} & \underline{0.432} & \underline{0.922} \\
\midrule
T5$_{base}$ & SFT & 0.840 & 0.874 & 0.832 & 0.775 & 0.830 & 0.430 & 0.912 \\
T5$_{base}$ & SFT+MDO$_{\text{min}}$ & \underline{0.872} & \underline{0.883} & \underline{0.850} & \underline{0.819} & \underline{0.856} & \underline{0.433} & \underline{0.918} \\
T5$_{base}$ & SFT+MDO$_{\text{pro}}$ & \textbf{0.882} & \textbf{0.887} & \textbf{0.858} & \textbf{0.836} & \textbf{0.866} & \textbf{0.435} & \textbf{0.922} \\
\midrule
\textcolor{gray}{GPT-4} & \textcolor{gray}{-} & \textcolor{gray}{0.973} & \textcolor{gray}{0.843} & \textcolor{gray}{0.831} & \textcolor{gray}{0.971} & \textcolor{gray}{0.904} & \textcolor{gray}{0.443} & \textcolor{gray}{0.851} \\
\bottomrule
\end{tabular}}

\caption{The results of automatic multi-dimension evaluation measured on the BillSum dataset. Within the same baseline, the bold denotes the highest score, and the underline denotes the second-highest score.}\label{tab:main}
\end{table*}



%% file: fig/tab-cnn.tex

\begin{table*}
\centering
\scalebox{0.78}
{\begin{tabular}{l|c|ccccc|cc}
\toprule
\multicolumn{2}{l|}{} & \multicolumn{5}{c|}{UniEval} & \multicolumn{2}{c}{} \\ 
\midrule
{Model} & {Fine-tune} & Coherence & Consistency & Fluency & Relevance & Overall & QuestEval & BERT{\small Score} \\ 
\midrule
PEGASUS & SFT & 0.936 & 0.939 & 0.815 & 0.684 & 0.843 & 0.584 & 0.877\\
BRIO  & SFT & 0.951 & 0.931 & 0.826 & 0.776 & 0.871 & 0.619 & 0.883 \\
\midrule
BART$_{base}$  & SFT & \textbf{0.963} & 0.952 & 0.850 & 0.702 & 0.867 & \textbf{0.594} & 0.877 \\ 
BART$_{base}$ & SFT+MDO$_{\text{min}}$ & 0.955 & \underline{0.958} & \underline{0.894} & \underline{0.734} & \underline{0.885} & 0.555 & \textbf{0.896} \\
BART$_{base}$ & SFT+MDO$_{\text{pro}}$ & \underline{0.959} & \textbf{0.960} & \textbf{0.896} & \textbf{0.750} & \textbf{0.891} & \underline{0.556} & \textbf{0.896} \\
\midrule
\textcolor{gray}{GPT-3+CoT} & \textcolor{gray}{-} & \textcolor{gray}{0.948} & \textcolor{gray}{0.870} & \textcolor{gray}{0.948} & \textcolor{gray}{0.910} & \textcolor{gray}{0.919} & \textcolor{gray}{0.574} & \textcolor{gray}{0.874} \\
\textcolor{gray}{GPT-4} & \textcolor{gray}{-} & \textcolor{gray}{0.967} & \textcolor{gray}{0.840} & \textcolor{gray}{0.945} & \textcolor{gray}{0.934} & \textcolor{gray}{0.921} & \textcolor{gray}{0.597} & \textcolor{gray}{0.864} \\
\bottomrule
\end{tabular}}

\caption{The results of automatic multi-dimension evaluation measured on the CNN/DailyMail (CNN/DM) dataset.}\label{tab:cnn}
\end{table*}




%% file: content/experiment.tex
\section{Experimental Setup}
\paragraph{Datasets} 

We utilize two text summarization datasets considering potential influences of source document complexity: the BillSum dataset for legislative content and the CNN/Daily Mail dataset for news summarization. BillSum comprises an 18.9K training set and a 3.2K test set, while CNN/DM has a 287K training set and an 11.5K test set. In light of studies indicating poor quality of reference summaries in the datasets \cite{liu2023learning, zhang2023benchmarking, goyal2023news}, we use an enhanced version of CNN/DM test set introduced by \citet{wang2023element}.

\paragraph{Baseline models} As baseline models, we employ encoder-decoder models commonly used for the text summarization task, including BART \cite{lewis-etal-2020-bart} and T5 \cite{2020t5}. 
For additional comparison, we report PEGASUS \cite{zhang2020pegasus} and BRIO \cite{liu-etal-2022-brio} results. To ensure comparability, we fine-tune BART{\small-base}, BART{\small-large}, and T5{\small-base} under the same hyperparameter settings: a batch size of 4, a learning rate of 5e-5, and 10 epochs. For PEGASUS and BRIO models, we utilized already fine-tuned versions on the Billsum\footnote{https://huggingface.co/google/pegasus-billsum} and CNN/DM\footnote{https://huggingface.co/google/pegasus-cnn\_dailymail}\footnote{https://huggingface.co/Yale-LILY/brio-cnndm-cased}.

In addition, we compare our model with LLMs, GPT-3-CoT \cite{wang2023element} and GPT-4 \cite{openai2024gpt4}.  GPT-3-CoT is a 2-stage chain-of-thought approach where the first stage extracts the core elements, and the second stage integrates them to address the issue of LLMs not sufficiently incorporating elements in generated summaries in the news datasets. We used \texttt{GPT-4-turbo} for GPT-4.

\paragraph{Hyperparameters for RL} For RL, we use a batch size of 4, a learning rate of 1.41e-6, discount factor $\gamma=0.9$, and randomly select only 10K samples from the training set of each dataset. We conduct experiments with three different seeds and report the average scores.

\input{fig/fig-real_example}

\input{fig/fig-chatgpt}

\paragraph{Metrics} We use various evaluation metrics for multi-dimension assessment, such as \texttt{UniEval}, ChatGPT, and human evaluations. For detailed measurements on each dimension, we also use \texttt{QuestEval} \cite{scialom2021questeval} and \texttt{BERTSCore} \cite{zhang2020bertscore}. \texttt{QuestEval} assesses precision by generating questions from summaries using a question generation model and checking if the answers are in the document. It generates questions from the document and verifies whether the answers are in the summaries for recall. The overall \texttt{QuestEval} score is an F1 score based on precision and recall. We use precision value for the \texttt{BERTScore}, which calculates the similarity between the token vectors in the generated summaries and those in the reference summaries based on BERT embeddings.

%% file: fig/fig-real_example.tex
\begin{figure*}[]
\centering
\includegraphics[width=0.95\textwidth]{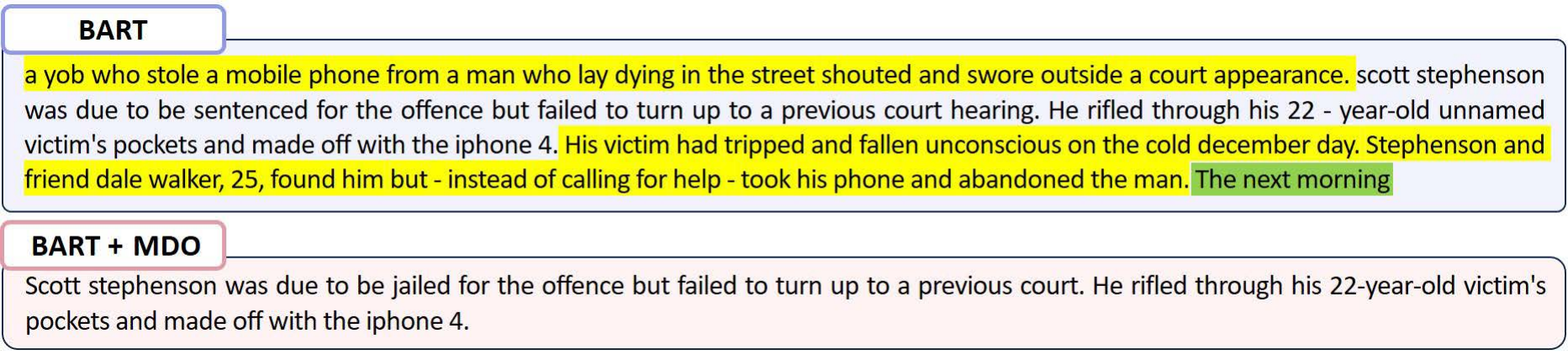}
\caption{Examples of the generated summaries by the baseline model and our MDO$_\text{pro}$ on the same document. Unimportant contents are highlighted in \colorbox{yellow!80}{yellow}, and unnatural or structurally disruptive ones are marked in \colorbox{green!70}{green}.}
\label{fig: real_example}
\end{figure*}

%% file: fig/fig-chatgpt.tex
\begin{figure}[]
\centering
\includegraphics[width=0.48\textwidth]{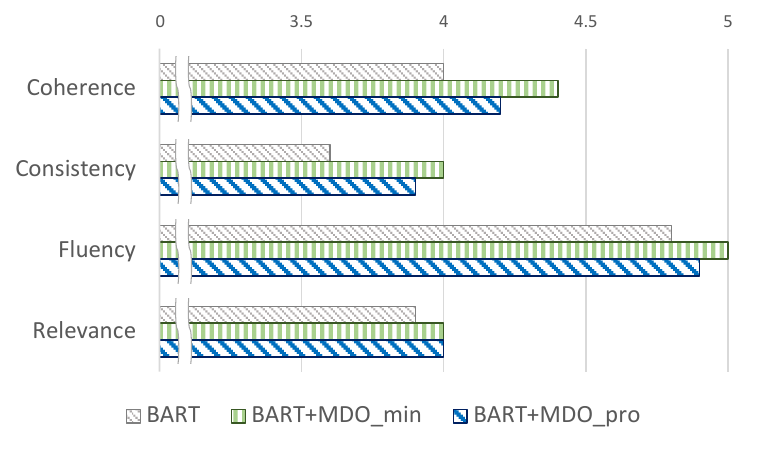}
\caption{Multi-dimensional evaluation results with ChatGPT on the BillSum.}
\label{fig: chatgpt}
\end{figure}

%% file: content/result.tex
\section{Results}

\paragraph{Main results} In Table \ref{tab:main}, our multi-objective optimization techniques, MDO${_\text{min}}$ and MDO${_\text{pro}}$, have consistently demonstrate enhanced performance across all \texttt{UniEval} dimensions. Notably, applying to the BART{\small-base} exhibit significant advancements in the lowest-quality dimension, \textit{relevance}, with MDO${_\text{min}}$ and MDO${_\text{pro}}$ showing increases of 0.027 and 0.034, respectively. Similarly, in the dimension of \textit{consistency}, also had inferior quality, MDO${_\text{min}}$ and MDO${_\text{pro}}$ lead to notable improvements of 0.024 and 0.020, respectively. Our methods consistently yield modest yet discernible enhancements even in dimensions with relatively high baseline scores. The same trend is evident in the evaluation of the BART{\small-large} model, with considerable strides made in dimensions that initially exhibited lower performance, accompanied by marginal but discernible improvements in dimensions already featuring high scores. This underlines adaptive learning capabilities our methods, enabling the model to dynamically adjust its focus and balance diverse dimensions with the overall enhancements. In the assessment using alternative metrics such as \texttt{QuestEval} and \texttt{BERTScore}, the BART{\small-large}+MDO$_{\text{min}}$ model stands out. These results highlight that our generated summaries maintain competitive quality even when measured based on the original document and the reference summaries. The standard deviation is specified in Appendix A.1.

We extend our experiments to include the CNN/DM dataset. As illustrated in Table \ref{tab:cnn}, training with multi-dimensional optimized methods enhances the performance on the CNN/DM dataset akin to those observed on the BillSum dataset. Notably, substantial score improvements are recorded in the dimensions of \textit{fluency} and \textit{relevance}, registering increases of 0.046 and 0.048, respectively, addressing areas where the quality was initially deficient. Still, the scores remained comparable or slightly lower in dimensions where the model already demonstrated high proficiency, such as \textit{coherence} and \textit{consistency}. Consequently, RL with MDO has resulted in well-balanced summaries across various dimensions.

As LLMs have demonstrated superior performance in summarization tasks \cite{zhang2023benchmarking, goyal2023news, pu2023summarization}, we compare our model with the latest LLMs, GPT-3+CoT \cite{wang2023element} and GPT-4 \cite{openai2024gpt4}. Despite the smaller model size, our method exhibits comparable performance to the larger and more expensive GPT-4 with only 0.018 differences in BillSum (Table \ref{tab:main}). Moreover, it shows higher \texttt{BERTScore} in BillSum and CNN/DM (Table \ref{tab:cnn}).

Figure \ref{fig: real_example} illustrates actual changes in summaries as multi-dimensional scores increase with our model. The initial models frequently incorporated irrelevant details and awkwardly constructed sentences. In contrast, our model, fine-tuned to enhance each dimension through MDO, effectively omits non-essential information and improves the natural flow of sentences. Qualitative observations suggest a positive link between improving \texttt{UniEval} scores and producing high-quality summaries.



\paragraph{ChatGPT evaluation} 
Recent studies informed that ChatGPT's evaluation capabilities closely align with human judgments \cite{gao2023humanlike, chiang-lee-2023-large, wang-etal-2023-chatgpt}. To further verify with indicators other than the QA-based metrics, we include ChatGPT evaluation with four dimensions identical to those in \texttt{UniEval}. Inputting the document and its summaries into ChatGPT, we request evaluations for each dimension on a scale ranging from 0 to 5 (the highest) using detailed prompts. As depicted in Figure \ref{fig: chatgpt}, the model with MDO$_{\text{min}}$ and MDO$_{\text{pro}}$ exhibits improvements across all evaluated dimensions compared to the baseline model, particularly demonstrating a noteworthy 11.1\% and 8.3\% increase in the lowest-rated dimension, \textit{consistency}. The prompts are shown in Appendix B.

%

\input{fig/tab-human}

\input{fig/fig-human_preference}

\paragraph{Human evaluation}
Given that the English-written BillSum dataset has congressional information, we hired three experts who are native English speakers and possess extensive experience with government documents via Upwork\footnote{https://www.upwork.com}. We follow the evaluation criteria outlined in \citet{roit-etal-2023-factually}, which employed NLI-based RL: \textit{comprehension}, \textit{attribution}, \textit{salience}, and \textit{conciseness}. \textit{comprehension} assesses the ease of understanding the summary, \textit{attribution} gauges the consistency of the summary with the main document, \textit{salience} determines whether the summary includes only the most important information, and \textit{conciseness} evaluates the brevity of the summary. As outlined in Table \ref{tab: human}, our model surpasses the baseline across all dimensions. Moreover, evaluators preferred summaries generated by our model over those produced by the baseline model, as depicted in Figure \ref{fig: human_preference}. To verify our methods, we conduct significance tests on the BillSum dataset for both human evaluation results and the \texttt{UniEval} \textit{overall} score.
The results of the two-tailed paired t-test, with p-values < 0.05, demonstrate statistically significant performance differences in MDO$_{\text{min}}$ and MDO$_{\text{pro}}$ compared to the baseline model, BART.


\input{fig/tab-mechanical}

\input{fig/tab-ablation}

\paragraph{Mechanical analysis} Recent studies \cite{liu2023learning, zhang2023benchmarking, goyal2023news} pointed out that reference summaries generally exhibit low quality; thus, ROUGE, which solely relies on overlap with reference summaries may not accurately capture the true quality of the summaries. Nevertheless, we assess our model using traditional evaluation metrics, including ROUGE, \texttt{coverage}, and the average summary length. Summaries of our model show relatively lower ROUGE scores (Table~\ref{tab: mechanical}), yet the comparative \texttt{coverage}, which calculates the proportion of tokens in the generated summary that are present in the document.

Meanwhile, models with MDO produce shorter summaries compared to those generated by base models. Comprehensive results of the substantial \texttt{coverage}, high \textit{relevance} and \textit{salience} scores (Table~\ref{tab:main}, \ref{tab: human}) imply that our shorter summaries concisely encapsulate only the essential contents from the document. In contrast, summaries generated by PEGASUS average around 193 words, which is excessively long for a summary. As demonstrated by \citet{guo-vosoughi-2023-length}, lengthy summaries are favorable in mechanical metrics like ROUGE. Further, \citet{roit-etal-2023-factually} reported a decrease in entailment percentage as the token length increases. The observed trends persist in our results, where PEGASUS, producing the longest summaries, shows the highest ROUGE scores.

%% file: fig/tab-human.tex
\begin{table}[]
\centering
\scalebox{0.71}{%
\begin{tabular}{l|cccc}
\toprule
    & \small{Comprehension} & \small{Attribution} & \small{Salience} & \small{Conciseness} \\ \midrule
BART & 4.11 & 3.81 & 3.81 & 3.76 \\
BART+MDO$_{\text{min}}$ & 4.73 & 4.17 & 4.36 & 4.74 \\
BART+MDO$_{\text{pro}}$ & 4.80 & 4.42 & 4.55 & 4.75 \\
\bottomrule
\end{tabular}%
} \caption{Human evaluation for the BillSum dataset. The scores are the average by three human expert.}\label{tab: human}
\end{table}



%% file: fig/fig-human_preference.tex
\begin{figure}[]
\centering
\includegraphics[width=0.43\textwidth]{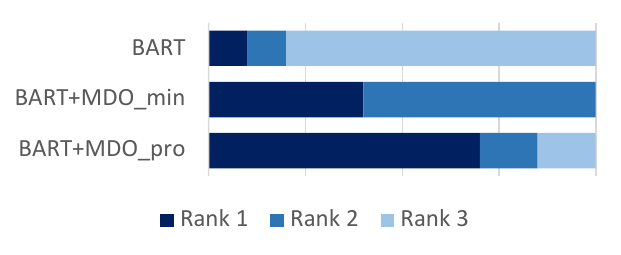}
\caption{Human preferences for each model. Rank 1 signifies the most preferred summary among the evaluated summaries.}
\label{fig: human_preference}
\end{figure}

%% file: fig/tab-mechanical.tex
\begin{table}[]
\centering
\scalebox{0.84}{%
\begin{tabular}{l|ccc}
\toprule
    & {\small{ROUGE-L}} & {\small{Coverage}} & {\small{Summary Length}} \\ \midrule
PEGASUS & 0.431 & 0.902 & 193.073 \\ 
\midrule
BART & 0.336  & 0.890 & 73.164 \\
BART+MDO$_{\text{min}}$ & 0.284 & 0.907 & 39.464 \\
BART+MDO$_{\text{pro}}$ & 0.276 & 0.898 & 37.002 \\
\midrule
T5  & 0.365 & 0.945 & 74.624 \\
T5+MDO$_{\text{min}}$  & 0.351 & 0.942 & 63.559 \\
T5+MDO$_{\text{pro}}$  & 0.340 & 0.939 & 55.957 \\
\bottomrule
\end{tabular}%
} \caption{The Mechanical Evaluation of summarization models. Our model generates brief summaries containing only the essential information.}\label{tab: mechanical}
\end{table}

%% file: fig/tab-ablation.tex
\begin{table*}
\centering
\scalebox{0.83}
{\begin{tabular}{l|c|ccccc|cc}
\toprule
\multicolumn{2}{l|}{} & \multicolumn{5}{c|}{UniEval} & \multicolumn{2}{c}{} \\ \cline{1-9}
 {Model} & {Fine-tune} & Coherence & Consistency & Fluency & Relevance & Overall & QuestEval & BERTScore \\ 
\midrule
BART$_{large}$ & SFT & 0.884 & 0.865 & 0.864 & 0.843 & 0.864 & 0.424 & 0.904 \\ 
\midrule
BART$_{large}$ & MDO$_{\text{sum-r}}$ & 0.922 & 0.931 & 0.465 & 0.916 & 0.809 & 0.448 & 0.929 \\ 
BART$_{large}$ & MDO$_{\text{sum-l}}$ & 0.892 & 0.887 & 0.872 & 0.861 & 0.878 & 0.431 & 0.924 \\
\midrule
BART$_{large}$ & MDO$_{\text{min}}$  &  0.899 & 0.894 & 0.882 & 0.869 & 0.886 & 0.435 & 0.924 \\
BART$_{large}$ & MDO$_{\text{pro}}$  & 0.900 & 0.895 & 0.877 & 0.871 & 0.886 & 0.432 & 0.922 \\
\bottomrule
\end{tabular}}


\caption{Comparison of performance between two naive methods of summing the rewards (MDO$_{\text{sum-r}}$) or losses (MDO$_{\text{sum-l}}$) and our two optimization methods (MDO$_{\text{min, pro}}$). Our strategies show better overall performance than the former two methods and show balanced results, unlike MDO$_{\text{sum-r}}$ exhibiting a severely low score for \textit{fluency}.}\label{tab:ablation}
\end{table*}


%% file: content/analysis.tex
\input{fig/fig-each_length}

\section{Discussions}
\paragraph{Summary length varies by text complexity} In text summarization tasks, concisely encapsulating only the critical information is crucial. However, the optimal length of a summary depends on the document's informational content, resulting in varying ideal lengths across datasets. When a document contains rich information, its summary tends to be longer; conversely, a document with less information leads to a shorter summary. The CNN/DM news dataset includes less information, allowing for the essential contents to be sufficiently covered in a shorter length. On the other hand, the legislative dataset, BillSum, characterized by longer texts and a greater volume of information, tends to yield longer summaries for all models, as revealed in Figure \ref{fig: each_length}. Remarkably, our models consistently produce short yet concise summaries for both datasets, while the PEGASUS model outputs severely lengthy summaries when the data complexity increases.


\input{fig/fig-length}

\paragraph{Discount factor affects summary length} We investigate the impact of a discount factor $\gamma$ on the length of the generated summaries. A clear pattern is found in our empirical experiments on the BillSum dataset -- a larger discount factor results in shorter summaries (see Figure \ref{fig: length}). This phenomenon can be attributed to the training process of the policy model, particularly its emphasis on the \textit{relevance} dimension. When estimating the advantage $A$, a larger $\gamma$ places more emphasis on future rewards. As the reward for the last token is determined using UniEval, and \textit{relevance} often receives the lowest score among dimensions, the training focus may lean heavily towards optimizing \textit{relevance}. Consequently, the model tends to anticipate higher scores by generating concise summaries that mostly include only the most crucial sentences, aligning with \textit{relevance}'s evaluation criteria of containing essential information. Thus, a larger discount factor is expected to generate shorter summaries in this specific context.



\paragraph{Comparison with naive approaches} 
When using RL in Language Models, careful attention should be paid to training, as models have the potential to diverge easily, and the value model may fail to converge properly. Considering the intricacy of multi-reward optimization, we conduct additional experiments, emphasizing the need for specialized optimization for multiple rewards. 
We explore straightforward optimization strategies, such as summing the rewards for each aspect score to formulate the final reward (MDO$_{\text{sum-r}}$) and aggregating the losses for each aspect score, akin to conducting multi-task training (MDO${_\text{sum-l}}$). However, employing the MDO$_{\text{sum-r}}$ method amplifies the performance gap between dimensions, making the superior ones better while the inferior ones (\textit{fluency}) worse, thereby boosting the imbalance. MDO$_{\text{sum-l}}$, a naive multi-task approach, shows improved results over the baseline but fails to outperform MDO$_{\text{min}}$ and MDO$_{\text{pro}}$ (Table \ref{tab:ablation}). These findings highlight the importance of our adaptive optimization strategies for multi-objective RL compared to simple multi-rewarding.



%% file: fig/fig-each_length.tex
\begin{figure}[]
\centering
\includegraphics[width=0.41\textwidth]{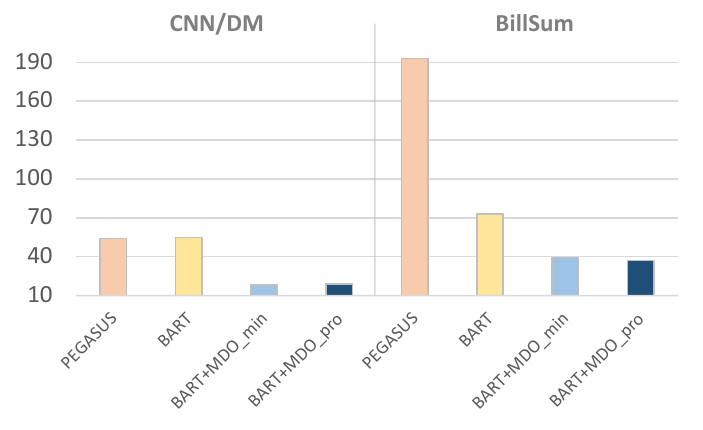}
\caption{Comparison of summary length for each model on different datasets. Even in complex data (BillSum; right), our methods produce shorter summaries.}
\label{fig: each_length}
\end{figure}




%% file: fig/fig-length.tex
\begin{figure}[]
\centering
\includegraphics[width=0.41\textwidth]{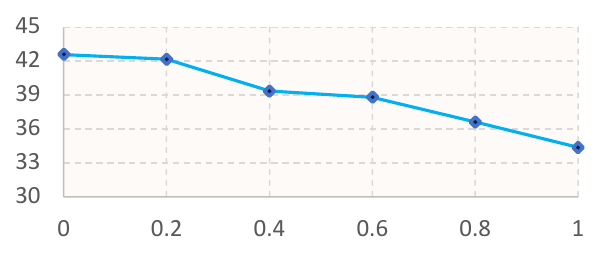}
\caption{Averaged length of generated summaries (y-axis) according to the discount factor $\gamma$ (x-axis).}
\label{fig: length}
\end{figure}



%% file: content/future.tex

%% file: content/conclusion.tex
\section{Conclusion}

This work aims to elevate the summary quality on diverse dimensions by introducing optimized multi-objective RL strategies. With the adoption of UniEval, we incorporate the assessed four-dimensional scores of summaries for rewarding. In particular, we propose two multi-dimensional optimization (MDO) strategies, aiming to learn the optimal policy during the multi-objective RL process. Our MDO strategies exhibited improved performance across all dimensions, and human-evaluated results further proved the capacity to generate balanced summaries. Comparisons with the naive summing of rewards or losses imply that our finer optimization strategies facilitates the efficacy of RL in summarization.

%% file: content/limitation.tex
\section*{Limitation}
In this work, we solely utilize UniEval, an open-source evaluation metric, for multi-dimensional evaluation due to its strong correlation with human judgment. However, our approach could be extended and applied if additional evaluation metrics for multiple dimensions become available. As a future work, combining multiple metrics for each single dimension can be further considered as in  \citet{wan2023faithfulnessaware}. We explored the relationship between the discount factor and summary length, yet did not investigate how it practically affects performance enhancement. Observing how performance varies by adjusting the discount factor could be an intriguing topic. Also, we employ MDO on the open-source small encoder-decoder models, considering their cost-effectiveness. This choice is attributed to our main goal of showcasing the applicability of multi-objective RL in summarization tasks. However, given the model-agnostic nature of MDO, implementation with other LLMs is feasible; thus, our method can be extended in future works.




%% file: content/ethics.tex
\section*{Ethical Statement}

We utilized public datasets such as BillSum, CNN/DM, and CNN/DM element-aware test sets in our research. For the human evaluation conducted through Upwork, we compensated fairly for the assessments. A total of \$50 was paid per person as a fixed prize for evaluating three summaries per document across ten documents, covering four dimensions and preference assessments.

%% file: content/acknowledge.tex
\section*{Acknowledgements}

This work was supported by the National Research Foundation of Korea (NRF) grant funded by the Korea government (MSIT) (No. RS-2023-00217286), Institute of Information \& communications Technology Planning \& Evaluation (IITP) grant funded by the Korea government (MSIT) (No.RS-2019-II191906, Artificial Intelligence Graduate School Program (POSTECH)), and the Technology Innovation Program (20015007, Development of Digital Therapeutics of Cognitive Behavioral Therapy for treating Panic Disorder) funded By the Ministry of Trade, Industry \& Energy (MOTIE, Korea).

%% file: content/appendix.tex
\appendix

\label{sec:appendix}

\section{Detailed Experimental Results}

\subsection{Standard deviation}
We evaluated the standard deviation for the experiments in Table \ref{tab:main} and Table \ref{tab:cnn}. The standard deviation results for each dataset are reported in Table \ref{tab:appendix-main-std} and Table \ref{tab:appendix-cnn-std}, respectively.

\input{fig/fig-appendix-main}

\input{fig/fig-appendix-cnndm}

\subsection{Performance variation according to the size of the value model}
We investigated whether the size of the policy and the value models influence the performance improvement extent in MDO. The UniEval, used as our reward, is based on the T5-large with 770M parameters. Compared to the reward model, the value models of BART-base (139M) and BART-large (406M) have smaller parameters. Consequently, it might be challenging for the value model to accurately predict rewards due to its relatively smaller size than the reward model. As shown in Figure \ref{fig: appendix-model_size}, the closer the value model's size to the reward model's size, the higher the performance improvement over the baseline.

\input{fig/fig-appendix-model_size}

\subsection{Performance differences based on the base optimizer of PCGrad}
In the MDO$_{\text{pro}}$, we utilized Adam as the base optimizer for PCGrad. The Adam optimizer adjusts the size of parameter updates based on the gradient magnitude, which results in significantly better performance compared to the SGD optimizer in the MDO$_{\text{pro}}$ method that involves gradient projection (Table \ref{tab: appendix-optimizer}).

\input{fig/tab-appendix-optimizer}

\subsection{Details of used metrics}

\begin{itemize}
    \item \texttt{UniEval} \cite{zhong-etal-2022-towards}: Evaluation model, which evaluates four dimensions with a single model. Each dimension is trained with questions and answers using T5. Scores for each dimension are calculated by inserting a prompt along with the summary.
    \item \texttt{QuestEval} \cite{scialom2021questeval}: Utilizes a question generation model to create questions from the document and checks if the answers to these questions are present in the summary, calculating recall. Conversely, it generates questions from the summary to check if the answers to these questions are present in the text, calculating precision.
    \item \texttt{BERTScore} \cite{zhang2020bertscore}: Calculates precision and recall through the cosine similarity between the token embeddings of the generated summary and the reference summary.
    \item \texttt{Coverage}: Measures whether each token of the generated summary is present in the document. Unlike exact copy, this metric is finely calculated through lemmatization and case conversion using the NLTK\footnote{https://www.nltk.org} library.
    \item ROUGE\footnote{https://huggingface.co/spaces/evaluate-metric/rouge}: Counts the number of overlapping words between the generated summary and the reference summary.
    \item Summary length: Counts the total word of the summary.

\end{itemize}

\subsection{Hardware usage}
For MDO, we used NVIDIA A100-SXM4-80GB, and for fine-tuning the baseline models on text summarization, we utilized NVIDIA RTX A5000.

\section{Detailed Evaluation Setup}

\subsection{ChatGPT evaluation}

For the ChatGPT\footnote{https://chat.openai.com} evaluation, we specified how each summary should be assessed. Providing a detailed description of the dimensions enables ChatGPT to assess each dimension properly. Scores were assigned on a scale from 0 to 5 (the highest) points. When given detailed prompts to evaluate each dimension, ChatGPT provides scores for each dimension along with explanations for its evaluations. For instance, if the summary includes incorrect information, such as hallucinations, ChatGPT will measure a low consistency score and provide an explanation for this assessment. The details of prompts are in Table \ref{tab: appendix-chat}.

\input{fig/tab-appendix-chatgpt-description}

\subsection{Human evaluation}

For our human evaluation, we hired three English-native experts through Upwork. We provided detailed scripts on how each dimension should be evaluated. Instead of using the dimensions of \textit{coherence}, \textit{consistency}, \textit{fluency}, and \textit{relevance} measured by \texttt{UniEval}, which we used as rewards, we followed the human evaluation dimensions used by \citet{roit-etal-2023-factually}. As the four dimensions used for our rewards are core elements in assessing the summary quality, we assumed that optimizing all four core elements would likely lead to positive evaluations in other unused dimensions as well. The detailed description we provided for human evaluation is illustrated in Table \ref{tab: appendix-human}.

\input{fig/tab-appendix-human_evalation_survey}

%% file: fig/fig-appendix-main.tex
 


\begin{table*}[!ht]
\centering
\scalebox{0.8}
{\begin{tabular}{l|c|ccccc|cc}
\hline
\multicolumn{2}{l|}{} & \multicolumn{5}{c|}{UniEval} & \multicolumn{2}{c}{} \\ \cline{1-9}
{Model} & {Fine-tune} & Coherence & Consistency & Fluency & Relevance & Overall & QuestEval & BERT{\small Score} \\ 

\hline
BART$_{base}$ & SFT+MDO$_{\text{min}}$ & $\pm{0.011}$ & $\pm{0.013}$ & $\pm{0.013}$ & $\pm{0.011}$ & $\pm{0.007}$ & $\pm{0.002}$ & $\pm{0.005}$ \\
BART$_{base}$ & SFT+MDO$_{\text{pro}}$  & $\pm{0.009}$ & $\pm{0.010}$ & $\pm{0.019}$ & $\pm{0.010}$ & $\pm{0.004}$ & $\pm{0.001}$ & $\pm{0.004}$ \\
\hline
BART$_{large}$ & SFT+MDO$_{\text{min}}$  & $\pm{0.002}$  & $\pm{0.001}$ & $\pm{0.022}$ & $\pm{0.003}$ & $\pm{0.006}$ & $\pm{0.001}$ & $\pm{0.006}$ \\
BART$_{large}$ & SFT+MDO$_{\text{pro}}$  & $\pm{0.007}$ & $\pm{0.005}$ & $\pm{0.008}$ & $\pm{0.006}$ & $\pm{0.003}$ & $\pm{0.002}$ & $\pm{0.005}$ \\
\hline
T5$_{base}$ & SFT+MDO$_{\text{min}}$ & $\pm{0.016}$ & $\pm{0.007}$ & $\pm{0.016}$ & $\pm{0.019}$ & $\pm{0.014}$ & $\pm{0.002}$ & $\pm{0.004}$ \\
T5$_{base}$ & SFT+MDO$_{\text{pro}}$ & $\pm{0.008}$ & $\pm{0.006}$ & $\pm{0.018}$ & $\pm{0.008}$ & $\pm{0.009}$ & $\pm{0.001}$ & $\pm{0.001}$ \\
\hline
\end{tabular}}

\caption{The standard deviation for the MDO$_{\text{min}}$ and MDO$_{\text{pro}}$ models in the BillSum dataset.}\label{tab:appendix-main-std}
\end{table*}



%% file: fig/fig-appendix-cnndm.tex

\begin{table*}
\centering
\scalebox{0.8}
{\begin{tabular}{l|c|ccccc|cc}
\hline
\multicolumn{2}{l|}{} & \multicolumn{5}{c|}{UniEval} & \multicolumn{2}{c}{} \\ \cline{1-9}
{Model} & {Fine-tune} & Coherence & Consistency & Fluency & Relevance & Overall & QuestEval & BERT{\small Score} \\ 
\hline

BART$_{base}$ & SFT+MDO$_{\text{min}}$ & $\pm{0.010}$ & $\pm{0.008}$ & $\pm{0.008}$ & $\pm{0.012}$ & $\pm{0.005}$ & $\pm{0.003}$ & $\pm{0.014}$ \\
BART$_{base}$ & SFT+MDO$_{\text{pro}}$ & $\pm{0.008}$ & $\pm{0.006}$ & $\pm{0.008}$ & $\pm{0.019}$ & $\pm{0.009}$ & $\pm{0.006}$ & $\pm{0.028}$ \\
\hline
\end{tabular}}

\caption{The standard deviation for the MDO$_{\text{min}}$ and MDO$_{\text{pro}}$ models in the CNN/DM dataset.}\label{tab:appendix-cnn-std}
\end{table*}



%% file: fig/fig-appendix-model_size.tex

\begin{figure*}[!ht] 
\centering
\includegraphics[width=0.7\textwidth]{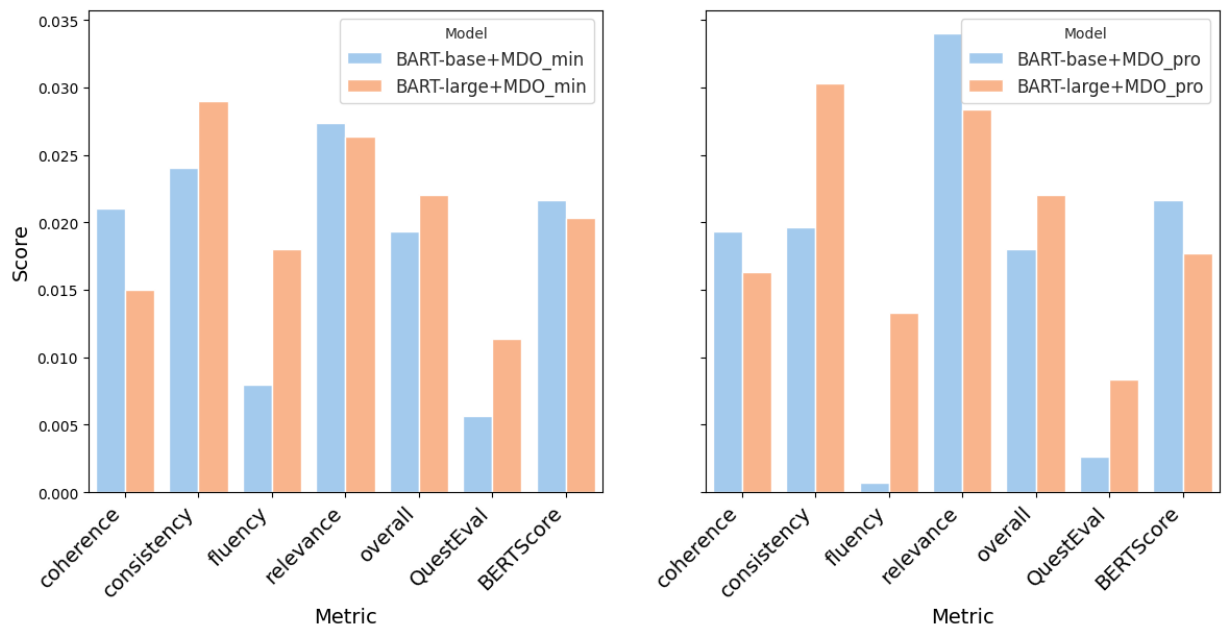}
\caption{Performance improvement degree over the baseline model according to the value model size.}
\label{fig: appendix-model_size}
\end{figure*}


%% file: fig/tab-appendix-optimizer.tex
\begin{table}[]
\centering
\scalebox{0.70}{%
\begin{tabular}{l|ccccc}
\hline
    & \small{Coherence} & \small{Consistency} & \small{Fluency} & \small{Relevance} & \small{Overall} \\ \hline
BART & 0.963 & 0.952 & 0.850 & 0.702 & 0.867 \\
\hline
MDO$_{\text{pro-SGD}}$ & 0.957 & 0.951 & 0.862 & 0.707 & 0.869 \\
MDO$_{\text{pro-Adam}}$ & 0.959 & 0.960 & 0.896 & 0.750 & 0.891 \\
\hline
\end{tabular}%
} \caption{In MDO$_{\text{pro}}$, the choice of the base optimizer for PCGrad leads to performance differences.}\label{tab: appendix-optimizer}
\end{table}

%% file: fig/tab-appendix-chatgpt-description.tex
\begin{table}[]
\centering
\scalebox{0.70}{%
\begin{tabular}{p{10cm}}
\hline
 \small{Description of the ChatGPT evaluation} \\ \hline

 Please evaluate the summaries. The dataset contains government and legislative data. Please evaluate three summaries per document on four aspects. The aspect required for the evaluation is as follows (score each aspect between 0 and 5, highest score of 5.0).
  \\ \\

 1. Coherence: Whether all the sentences form a coherent body.
 
 2. Consistency: Factual alignment between the summary and the source document.
 
 3. Fluency: The quality of individual sentences.
 
 4. Relevance: Whether the summary contains only the important information of the source document. \\

\hline
\end{tabular}%
} \caption{}\label{tab: appendix-chat}
\end{table}

%% file: fig/tab-appendix-human_evalation_survey.tex
\begin{table}[]
\centering
\scalebox{0.70}{%
\begin{tabular}{p{10cm}}
\hline
 \small{Description of the human evaluation} \\ \hline

 Please Evaluate the summaries. The dataset contains government and legislative data. Please evaluate three summaries per document on four aspects. The aspect required for the evaluation is as follows (score each aspect between 0 and 5, highest score of 5.0) \\ \\

 1. Comprehension: Is that summary easy to understand?
 
 2. Attribution: Is that summary consistent with the document?
 
 3. Salience: Does that summary contain only important information? (There should be no unimportant content)
 
 4. Conciseness: Is that summary short enough as a summary?
 
 5. Overall: The overall score of the summary (in your preferences). \\

\hline
\end{tabular}%
}\caption{} \label{tab: appendix-human}
\end{table}